%% file: main.tex
\documentclass[10pt]{article}
\usepackage{amsmath}
\usepackage{graphicx}
\usepackage{amsfonts}
\usepackage{hyperref}
\usepackage{tikz}
\usepackage{pdfpages}
\usetikzlibrary{positioning}
\usetikzlibrary{fit}

\begin{document}
    \includepdf{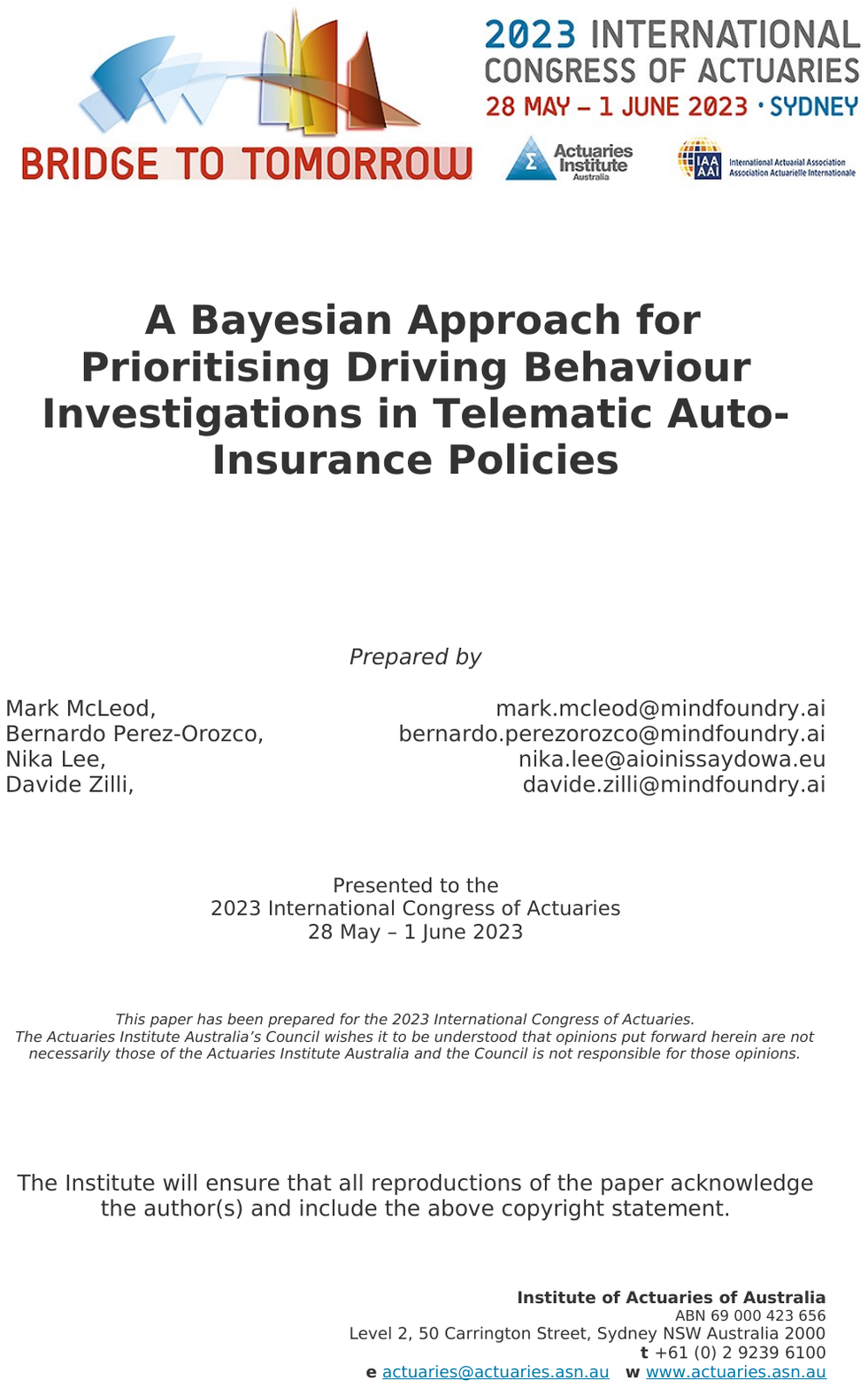}
    \begin{abstract}
        In this work, we use a probabilistic machine learning approach to efficiently identify drivers who may be using their vehicles for commercial purposes.

        Automotive insurers increasingly have access to telematic information via black-box recorders installed in the insured vehicle, and wish to identify undesirable behaviour which may signify increased risk or uninsured activities.
        However, identification of such behaviour with machine learning is non-trivial, and results are far from perfect, requiring human investigation to verify suspected cases.
        An appropriately formed priority score, generated by automated analysis of GPS data, allows underwriters to make more efficient use of their time, improving detection of the behaviour under investigation.

        An example of such behaviour is the use of a privately insured vehicle for commercial purposes, such as delivering meals and parcels.
        We first make use of trip GPS and accelerometer data, augmented by geospatial information, to train an imperfect classifier for delivery driving on a per-trip basis.
        We make use of a mixture of Beta-Binomial distributions to model the propensity of a policyholder to undertake trips which result in a positive classification as being drawn from either a rare high-scoring or common low-scoring group, and learn the parameters of this model using Markov-chain Monte-Carlo (MCMC).
        This model provides us with a posterior probability that any policyholder will be a regular generator of automated alerts (drawn from the higher-scoring group and therefore likely to be engaging in delivery driving) given any number of trips and alerts.
        This posterior probability is converted to a priority score, which was used to select the most valuable candidates for manual investigation.

        Testing over a 1-year period ranked policyholders by likelihood of commercial driving activity on a weekly basis.
        The top ~0.9\% have been reviewed at least once by the underwriters at the time of writing, and of those 99.4\% have been confirmed as correctly identified, showing the approach has achieved a significant improvement in efficiency of human resource allocation compared to manual searching.

    \end{abstract}

    \section{Introduction}\label{sec:introduction}
    \input{src/introduction}

    \section{GPS Path Classification}\label{sec:gps-path-classification}
    \input{src/gps-path-classification}

    \section{Policy Classification}\label{sec:policy-classification}
    \input{src/policy-classification}

    \section{Implementation}\label{sec:implementation}
\begin{figure}
        \centering
        \includegraphics[width=0.85\columnwidth]{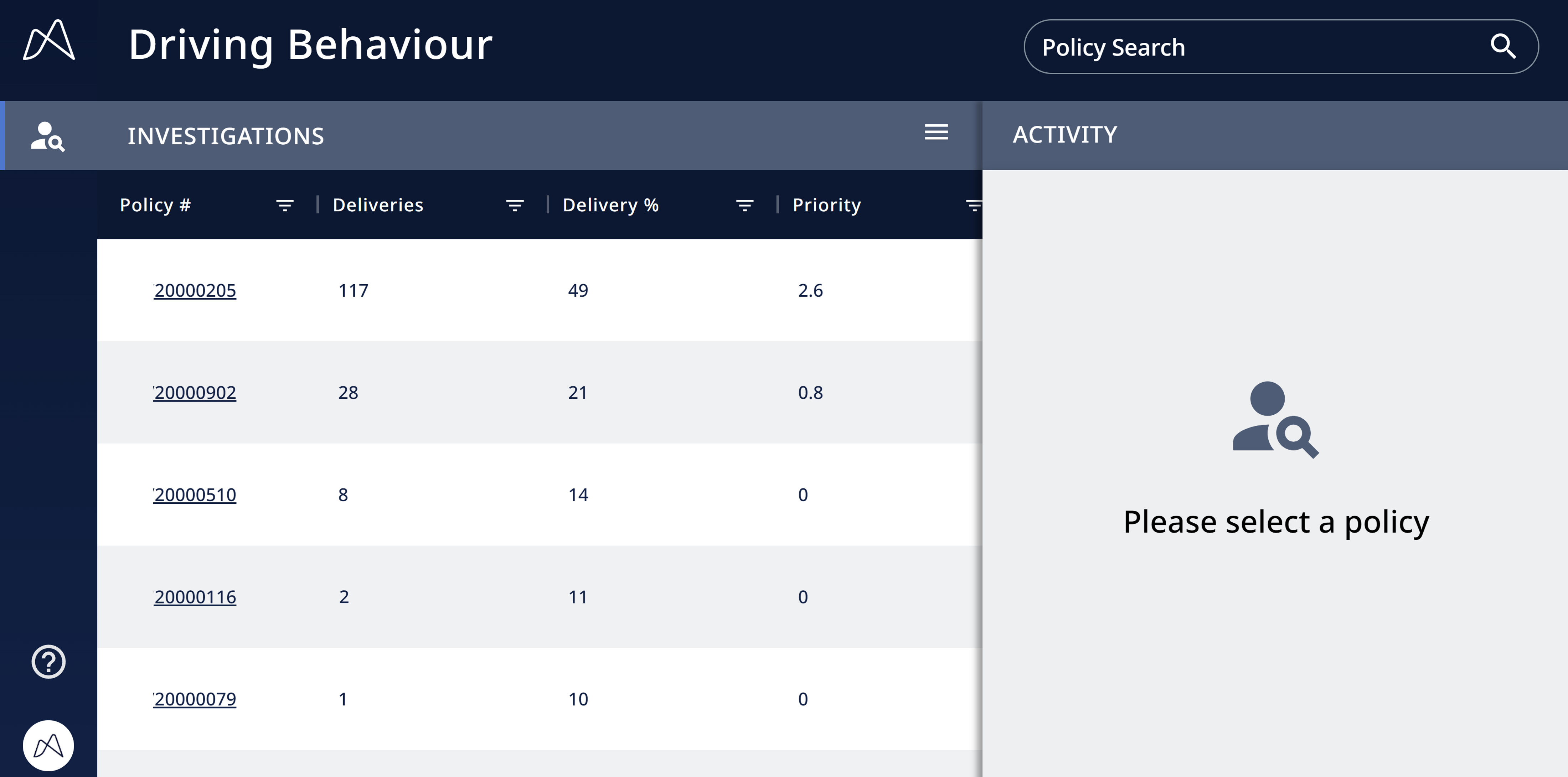}
        \caption{Screenshot of web application (on dummy data) developed to review drivers suspected to carry out commercial activities outside their policy T\&Cs. The Priority column provides the output of the model described in Section~\ref{sec:policy-classification}}
        \label{fig:screenshot}
    \end{figure}

    Following the construction of both a trip-level GPS path classifier, and the driver-level and policy-level classifiers, we developed a web application to allow underwriters to continuously validate the results produced.
    The pipeline described in Sections~\ref{sec:gps-path-classification} and ~\ref{sec:policy-classification} produces new predictions at a regular cadence given updated data, which are then sorted by the resulting score, as shown in Figure~\ref{fig:screenshot}.

    Results are sorted by priority score, and underwriters can then review individual drivers top-bottom starting with those likeliest to be delivery drivers, thus maximising the meaningfulness of the limited time they can spend on manual reviews.

    \section{Results}\label{sec:results}
    \input{src/results}

    \section{Conclusions \& Future Work}\label{sec:conclusions-&-future-work}
    \input{src/conclusions-and-future-work}

    \bibliography{main}
    \bibliographystyle{ieeetr}

\end{document}

%% file: src/introduction.tex
Use of telematics data in auto-insurance is growing rapidly as new applications are developed to take the advantage of modern telematics systems and the widespread use of mobile devices.
At present, telematics insurance includes safe driving and mileage based products where product features such as coverage, service or pricing are based on the telematics data.

Common data captured by telematics systems includes location, speed, acceleration or braking, fuel consumption, idling time and vehicle faults.
This information can provide in-depth insights and be analysed for events and patterns.
In this paper we consider the particular use of vehicles for deliveries.

Delivery driving is considered business use and not usually covered by standard Social Domestic and Pleasure (SD\&P) personal auto-insurance.
This could mean that an individual driving for a company like Deliveroo, JustEat, UberEats or Amazon involved in an accident could find their insurer refusing to pay out, leaving the driver in a potentially devastating financial situation.
It also means they might have been driving illegally.
There is a product available called `top-up' Hire and Reward (H\&R) insurance, which a driver can buy as supplemental cover in parallel with their SD\&P insurance.
In this work we aim to efficiently identify delivery driving and so prevent scenarios where consumers may experience unsatisfactory outcomes as a result of misrepresentation or underinsurance.

Identification of policies undertaking delivery driving represents a twofold challenge: classification of individual journeys as being potentially being deliveries, and subsequent identification of policies for which to make an intervention given results for individual trips.
Both stages will provide significantly imperfect results, and contacting a customer based on a false positive has a significant impact, both in resources and reputation.
For this reason we require human investigation prior to any action.
However, the volume of data generated by telematics insurance is too great for manual review of all trips to be feasible.
In this study, we investigated the we make use of machine learning to provide a prioritised ordering of policies for human investigation.

The first stage of our solution uses an automated machine learning pipeline to provide classification of delivery driving on a per-trip basis, based on GPS data.
In isolation as an individual classifier, this model has a significant false positive rate, which combined with the highly imbalanced dataset dilutes the minority of correctly identified trips beyond practical investigation.
The second stage uses a Bayesian mixture model approach to identify two populations within all policyholders: a majority group with a low rate of trips being given positive classifications, and a minority group with a much higher rate of positive classification.
Policyholders can then be assigned a probability of membership of the minority (higher rate) class at any point in time based on the available data.
This Bayesian approach ensures that the volume of evidence available is appropriately balanced with the simple number, or fraction, of identified trips when providing a ranking.
Policyholders with all trips undertaken classified as delivery driving, but only one or two total trips counted, are still ranked lower than those with a lower fraction, but greater total number trips.
The resulting scores are used to propose a list of policyholders to investigators ordered by ranking.
This allocates time to only those policies that have been identified as having a high probability of being delivery drivers.

Considering these predictions over a period of 1 year, updated on a weekly basis, resulted in 99.4\% of policies for which manual investigation was undertaken being labelled as undertaking delivery driving (approximately 0.9\% of all unique policies present in the test period).
In Section~\ref{sec:gps-path-classification} we describe the data available and our approach to classifying individual trips as undertaking delivery driving.
We then describe our method for ranking policies for investigation based on individual trip classification results in Section~\ref{sec:policy-classification}, and give a brief description of how our methods were deployed for use by underwriters in Section~\ref{sec:implementation}.
We present results in Section~\ref{sec:results}, and discuss our conclusions, business benefits and possible further extensions in Section~\ref{sec:conclusions-&-future-work}.

%% file: src/gps-path-classification.tex
As described above, identifying delivery driver behaviour is a twofold challenge.
This section tackles the first stage, which requires classifying individual trips as DELIVERY or NON-DELIVERY.
Once a prediction has been made for each individual trip, results can be aggregated at the driver level.

\subsection{Data}\label{subsec:data}
The data used in this study is from a UK provider of telematic insurance with tens of millions of trips and hundreds of thousands of drivers available for modelling purposes.
Telematic data is arranged into a series of regularly sampled and timestamped GPS measurements, each with the following attributes:
\begin{itemize}
    \item Driver identifier
    \item Trip identifier
    \item Latitude, Longitude
    \item Date and time
    \item Engine status (on/off/running)
    \item Accelerometer
\end{itemize}
We note that raw data collected is post-processed to align with the UK road network with high accuracy.
Technical details on how to achieve this lie outside the scope of this paper and are omitted to comply with intellectual property protection.
In addition to the individual GPS measurements, postprocessing also identifies the start and endpoints for individual trips.
The definition of what constitutes a trip may vary depending on the application.
In the case of detecting delivery driving behaviour, we used a dividing window of 90 idle minutes.
This allows for idle time between deliveries, including time waiting for delivery preparation at commercial stops to not constitute a break into separate trips.

Working with this dataset poses several difficulties compared to a standard machine learning setting:
\begin{description}
    \item [Lack of labelled trip data] We note that trips do not come with a DELIVERY or NOT A DELIVERY label, which complicates building a supervised classifier.
    Although out-of-policy use for commercial purposes is known to be a significant issue anecdotally, a consistent set of labels did not exist.
    \item [Scarce and noisy policy-level labels] Only six occasional delivery drivers had been identified at the beginning of this project, with their trip data having mixed delivery and non-delivery usage.
    \item [Dataset size] Due to the number of trip and driver data points, exhaustive manual investigation whether by underwriting experts or data scientists was unfeasible.
    \item [Few previously known examples] Only a handful of drivers ($<$10) were previously confirmed to engage in delivery driving mixed with personal use.
    Furthermore, trip-level labels for these drivers were also unavailable.
    \item [Difficulty to compute quantitative metrics] The lack of labelled data also made it difficult to compute traditional classification metrics, such as precision and recall, for evaluation purposes.
\end{description}

\begin{figure}
    \centering
    \includegraphics[width=0.8\columnwidth]{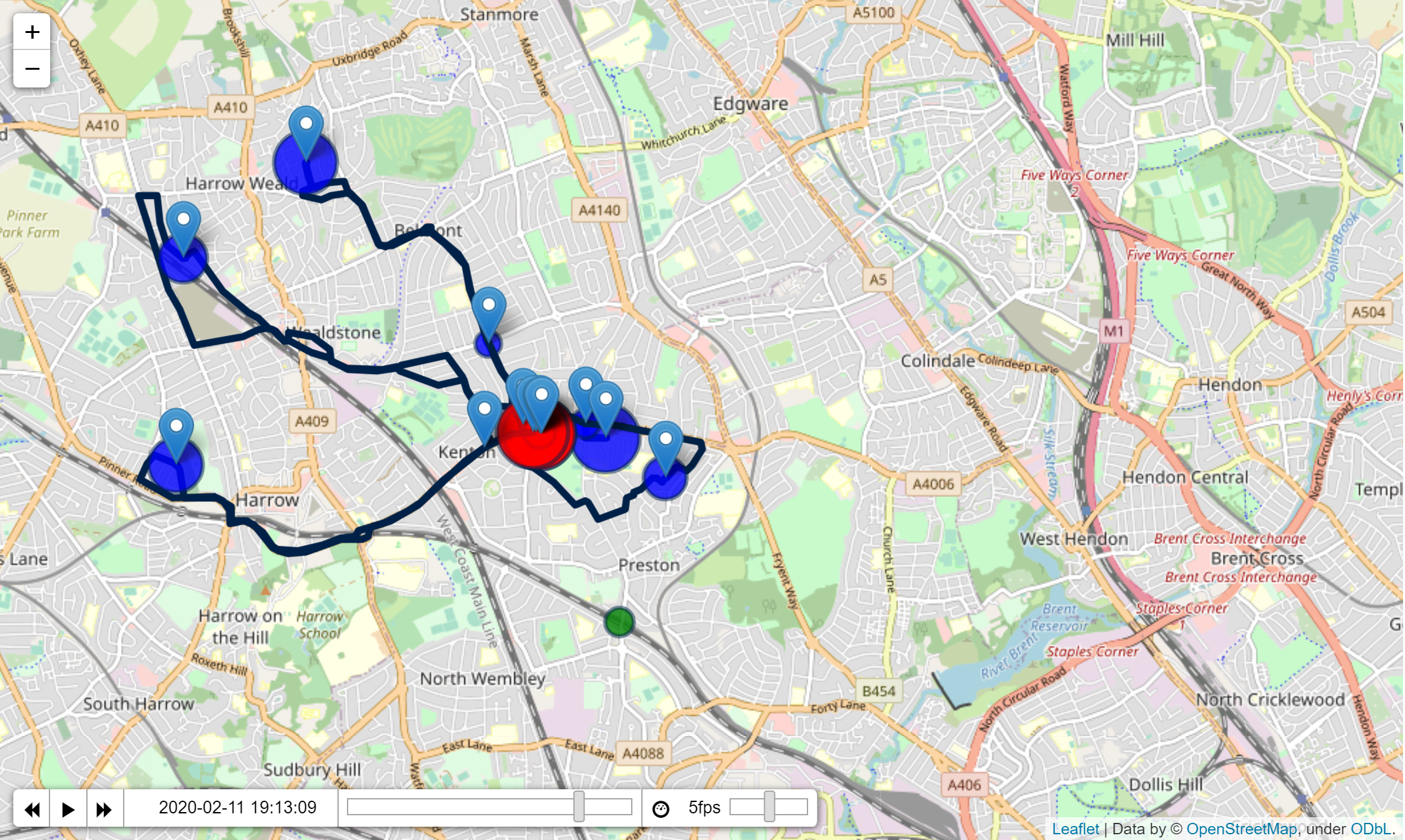}
    \caption{An example of a delivery trip candidate. Red markers represent stops at busy streets with nearby restaurants; blue markers represent stops at residential roads; and the green marker represents an (anonymised) policyholder home address. The radius of the marker is proportional to the amount of time they stayed there. The driver alternates between the blue markers, which they visit only for a few minutes before returning to the red markers area. This is a recognised pattern in delivery driving behaviour.}
    \label{fig:example-gps-trace}
\end{figure}

\subsection{Feature Engineering}\label{subsec:feature-engineering}
GPS trip data is of variable length, and much of the useful information is in relation to features in the world at or near to the given location, not inherent in the actual values.
To extract useful information which is suitable to be passed to a machine learning algorithm we must use domain knowledge to construct a feature set from the raw data.
The feature set we constructed aggregates trip-level features with a strong emphasis on qualifying the type and number of destinations in a single trip.
In doing so, we encapsulate underwriting domain knowledge concerning the behaviour of delivery drivers in real life.
For example, it is likely that delivery drivers perform multiple deliveries in one journey, which would suggest long trips with an alternation between residential and high-street-like destinations.
We provide an example of this behaviour in Figure~\ref{fig:example-gps-trace}.
Additionally, other trip-level characteristics such as time of the day, and day of the week, are also likely to inform the model.

We identified trip destinations by selecting consecutive trip stationary points that had an inter-sample absolute displacement of $\textrm{EPS} \leq 10^{-5}$.
Whenever a stationary point’s duration was shorter than 90 seconds it was discarded on the assumption that the driver may be stopped at a traffic light or other short term obstruction, rather than at a genuine trip destination.
Each stationary point was augmented using a database of commercial points of interest, such that:
\begin{itemize}
    \item a stationary point with 2 or more commercial locations within a radius of 50m was marked as a commercial destination;
    \item a stationary point within 150m of the driver’s home postcode was marked as a home destination; and
    \item a stationary point that does not satisfy either of the conditions above was marked as a residential destination.
\end{itemize}
Following the computation of trip stationary points, we include in our feature set:
\begin{description}
    \item  [\texttt{TRIP\_DURATION\_MINUTES}] Total trip duration in minutes.
    \item [\texttt{NUMBER\_WAITS\_TRIP}] Total number of stationary points in the trip.
   \item [\texttt{AVERAGE\_TRIP\_WAIT\_MINUTES}] Average duration of stationary points in trip, in minutes.
    \item [\texttt{TOTAL\_COMMERCIAL\_WAITS}] Number of commercial destinations.
    \item [\texttt{RATIO\_BUSY\_WAITS}] Ratio between number of commercial and residential destinations.
    \item [\texttt{TIME\_OF\_DAY\_TRIG}] Time of the day encoded through a trigonometric transformation.
\end{description}

\subsection{Modelling}\label{subsec:approach}
The remainder of this section explains how we constructed a supervised trip-level classifier.
First, we used an unsupervised approach, to guide identification of delivery trips from a limited set of policies which were known in advance to be undertaking delivery driving.
This allowed for a feasible, manual investigation by the insurer’s underwriting team to confirm trip-level labels, enabling a subsequent supervised classification stage.

\subsubsection*{Unsupervised delivery trip shortlisting}\label{subsubsec:unsupervised-delivery-trip-shortlisting}
Given the unlabelled nature of this telematic dataset, constructing a supervised classification model was not immediately feasible.
In order to curate an initial set of labels, we developed an unsupervised pipeline by exploiting a set of six confirmed delivery drivers.
These drivers had mixed usage, with some of their trips not being used for delivery purposes.

In order to create an initial shortlist of delivery trips, we constructed an unsupervised pipeline through of Uniform Manifold Approximation and Projection for dimension reduction (UMAP)~\cite{mcinnes2018umap} followed by Hierarchical Density-Based Spatial Clustering of Applications with Noise (HDBSCAN)~\cite{mcinnes2017hdbscan}, which was applied following the feature engineering pipeline previously described.
UMAP is a non-linear dimensionality reduction technique that more accurately preserves global and local structure, as well as non-linear relationships, than other approaches such as Principal Component Analysis or t-distributed Stochastic Neighbour Embedding.
This makes it suitable for being used in tandem with clustering approaches.
HDBSCAN is an extension of the density-based DBSCAN approach to clustering, which does not require tuning the number of sought clusters (unlike classical methods such as k-means clustering) this is advantageous for us as there may be an unknown number of characteristic driving behaviour patterns in addition to delivery driving.

\begin{figure}
    \centering
    \includegraphics[width=0.8\columnwidth]{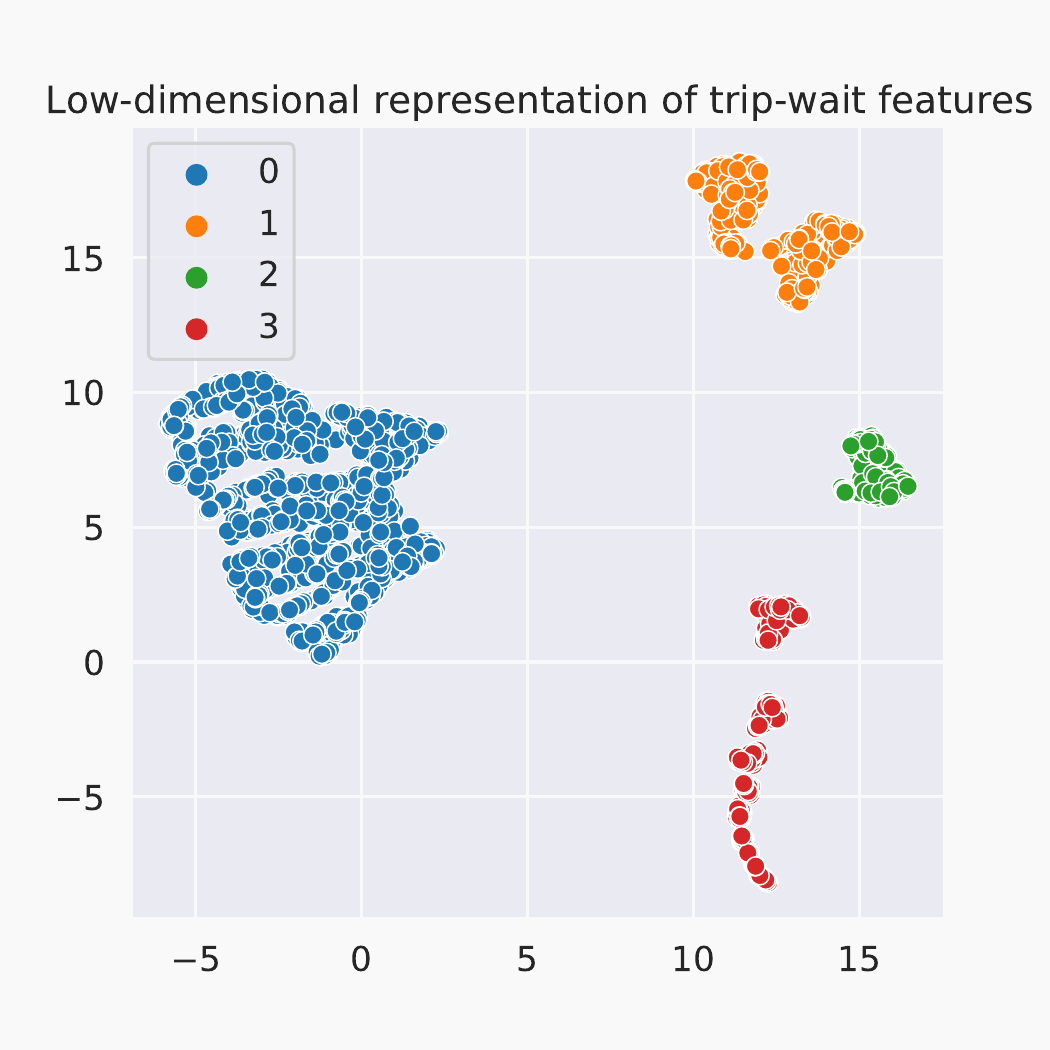}
    \caption{Output of our unsupervised pipeline on the deliveries dataset. This 2-dimensional representation of the dataset exhibits four distinct groups discovered by our approach. By matching up a handful of manually-identified true delivery trips with the clusters, we found that Cluster 3 (red) contained all of these examples. By submitting a random selection of 62 samples from Cluster 3 for manual review by the insurer's underwriting team, 58 of them were found to be true delivery trips.}
    \label{fig:umap}
\end{figure}
Running this clustering pipeline on all available trip data from the six confirmed delivery drivers resulted in a set of clusters as shown in Figure~\ref{fig:umap}, with one of them containing most of the trips with a high number of commercial destinations.
This cluster contained 62 candidate trips, which were manually reviewed by the insurer’s own underwriting team.
Following this review 58 out of the 62 trips were confirmed to be delivery trips.

\subsubsection*{Supervised classification}\label{subsubsec:supervised-classification}
Following the manual review of true delivery trip labels by the insurer's Underwriting team, these were used to construct a supervised model.
This was achieved through Mind Foundry Analyze, which is a workbench for automatically building and selecting from a wide range of models, including Extremely Randomised Forests, XGBoost Classifiers.
In addition to model training, MF Analyze is powered by its in-house Bayesian optimiser, which also finds the best model family with its corresponding optimal hyperparameters.
The resulting model is a trip-level classifier that takes as input the feature set described above, and returns a binary prediction.

\subsection{Conclusion}\label{subsec:conclusion-and-remarks}
This section introduced our methodology to develop a GPS path classifier for detecting delivery trips.
This was achieved by crafting a domain-driven feature set and making use of a one-off experiment to curate a set of high-confidence labels in collaboration with the insurer’s Underwriting department.
These labels were subsequently used to train a trip-level binary classifier on Mind Foundry’s end-to-end model building workbench, MF Analyze.

However, we note that the ultimate goal is to produce driver-level referrals, so we require an additional stage which aggregates trip-level predictions in a robust way to prioritise policies which are most likely undertaking delivery driving.
Simple approaches such as taking the simple delivery trip rate or raw count are unsuitable:
Using the raw count of delivery trips per policy will not to prioritise policies which have a high proportion of predicted delivery trips, but low total overall driving volume.
On the other hand, using the delivery trip rate, i.e.\ the ratio of delivery trips over the total number, will not prioritise policies that have a large number of delivery trips within an even higher total volume.
Given that the false positive rate of our model is not insignificant, and that delivery driving is considered to be a rare activity, we can anticipate that the majority of highly scored policies under a total rate metric would be those that are not undertaking delivery driving, but have all of only a small number of trips positively misclassified.
Furthermore, it must be noted that trip-level errors are likely to be biased by the behavior of a policyholder.
A driving pattern that the model misclassifies as a delivery, is likely to be regularly repeated by the same policyholder, leading to a much higher false positive rate for some policyholders than would be expected for independent trip-level classification.

The next section proposes a novel Bayesian approach to produce driver-level scores that prioritise the drivers that are most likely to be delivery drivers.
Our approach combines the counts of delivery and non-delivery classifications per policy to train a probabilistic mixture model based on the Beta-binomial distribution.
This provides a principled posterior probability of any individual policy undertaking delivery driving given the observed trip classifications.
Furthermore, by using a Bayesian approach, we are able to encapsulate Underwriting domain expertise in the shape of a prior distribution.

%% file: src/policy-classification.tex
We have described the provision of a per-trip classifier.
However, even with a reasonably high model performance, a non-negligible false positive rate and heavily imbalanced data will still most likely lead to an excess of positively classified trips, the majority of which are false positives.
Fortunately, we do not need to consider trips in isolation.

We hypothesize that trips are not incorrectly classified at random, but instead that policyholders may undertake particular non-delivery driving patterns which our model is unable to distinguish from delivery driving, and that the driving habits of a policyholder over several trips will have a particular expected rate of generating positive trip classifications.
However, policyholders actually undertaking delivery driving are expected to incur a much higher rate of positive classification.
The policyholder classification challenge is to identify these particular policies given the number of positive and negatively classified trips for each policyholder in a given interval.
Intuitively we seek some ranking methodology which will prioritise a high fraction of positively classified trips but discount a small total number of trips.
The highest scored policies from this secondary classification are then the most promising candidates for further investigation.

The output of the trip-classification process is a set of binary labels $\textrm{trip}_{ij} \in \{0, 1\}$ for trip $j$ by policyholder $i$.
Given this data we now wish to determine the posterior probability of an unseen label $k_i$, the likelihood that policyholder $i$ is part of a minority (delivery-driving) group  ( $k_i = 1$ ) which has a much higher rate of positive trip classifications than the larger population ( $k_i = 0$ )
We do not in this work use the ordering of trips, so this can be simplified to use only the count of trip and positive trip classifications
\begin{equation}
    p(k_i \mid \textrm{trip}_{i0}, \textrm{trip}_{i1}, \dots \textrm{trip}_{in})  = p(k_i \mid y_i,x_i)\label{eq:pdeliverygiventrips}
\end{equation}
where $x_i$ is the total number of trips, and $y_i$ is the number of positively labeled trips by policyholder $i$.
We are able to compute the number of trips $X={x_0, x_1, \dots x_N}$ and positively classified trips $Y={y_0, y_1, \dots y_N}$.
However, we have true labels $k_i$ for only an insignificant number of cases, so are not able to train a traditional supervised model to make predictions based on $p(k_i \mid y_i,x_i, X, Y)$.

To solve this problem of learning a mapping from classified trip counts to a class probability in the absence of labels, we choose to use our knowledge of the process generating the data to construct a probabilistic model which describes how delivery driving influences the classification of trips.
We can then estimate the unknown parameters, $\Theta$ of this model using the available data, and given these learned parameters provide a posterior probability of delivery driving given observed trip counts for each policy $ p(k_i \mid  x_i, y_i, \Theta)$.
We have some intuition for the likely values of $\Theta$, but do not know exact values, so we encode our uncertainty by choosing a set of hyperparameters $\Phi$, to impose a prior over the values of $\Theta$.

\begin{figure}
    \begin{centering}
        \begin{tikzpicture}[
        opennode/.style={circle, draw=black, fill=gray!0, very thick, minimum size=12mm},
        fillednode/.style={circle, draw=black, fill=gray!60, very thick, minimum size=12mm},
        platenode/.style={rectangle, draw=black, thick, minimum size=15mm, rounded corners},
        line/.style={very thick, shorten >= 5pt, },
        ]
%Nodes

        \node[opennode] at (3,0)     (theta)                           {$\theta$};
        \node[opennode] at (3,3)     (mode-params)                             {$\alpha_c, \beta_c$};
        \node[fillednode]  at (0,3)    (hyperparameters)                           {$\Psi$};
        \node[opennode]  at (6,0)      (k)        {$k_i$};
        \node[opennode]  at (6,3)      (q)        {$q_i$};
        \node[fillednode] at (9,3)       (x)        {$x_i$};
        \node[fillednode] at (9,0)       (y)        {$y_i$};
        \node[platenode,fit=(mode-params), label={below:$k \in \{0,1\}$}] {};
        \node[platenode,fit=(x)(y)(k)(q), label={below:$i \in 1,2,\dots, N$}] {};
%Lines
        \draw[-stealth] (hyperparameters) -- (mode-params);
        \draw[-stealth] (hyperparameters) -- (theta);
        \draw[-stealth] (theta) -- (k);
        \draw[-stealth] (k) -- (q);
        \draw[-stealth] (mode-params) -- (q);
        \draw[-stealth] (q) -- (y);
        \draw[-stealth] (x) -- (y);

        \end{tikzpicture}
        \caption{Directed Graphical Model for the latent variable mixture model construction. Only the data ${x, y}$ and the model hyperparameters $\Psi$ are observed, all other quantities must be inferred.}
        \label{fig:model_dag}
    \end{centering}
\end{figure}

We therefore wish to perform inference to learn the posterior distribution over the model parameters $p(\Theta \mid X, Y, \Phi)$ given the observed data and our hyperparameters.
We can then make predictions over the unobserved $k_i$ conditioned on this learned distribution
\begin{equation}
p(k_i \mid x_i, y_i, X, Y, \Phi) = \int  p(k_i \mid x_i, y_i, \Theta, \Phi)p(\Theta \mid X, Y, \Phi) \mathrm{d}\Theta.\label{eq:integralprob}
\end{equation}
Note that in this construction we have no information about unobserved class labels $k$.
Ensuring that it will represent the high-trip-classification-rate delivery-driving class we are interested in is determined by our model construction and choice of hyperparameters guiding the learning process.
In future iterations of this work, when significant numbers of verified observations of $k_i$ will be available, the model training process could be modified appropriately to accommodate this information.

Our choice for how to parameterize the quantities of interest and their dependencies determines the difficulty of the inference problem, and how likely we are to be successful.
As noted we are not considering the relative ordering of positively and negatively classified trips as features.
We also do not expect that the total number of trips made, $x_i$, is directly indicative of delivery driving, although it does represent the volume of available information, so does appear in modelling.
Given these assumptions the expected value of $y_i$ given $x_i$ for a policy can be expressed as
\begin{equation}
\mathbb{E}[ y_i \mid x_i, q_i] = x_i q_i, \qquad 0 \leq q_i \leq 1.
\label{eq:meany}
\end{equation}
The per-policy rate, $q_i$, is drawn from some unknown distribution which is not equal for the delivery-driving and non-delivery driving groups
\begin{equation}
p(q_i \mid k_i = 0)  \neq p(q_i | k_i = 1). \label{eq:classesunequal}
\end{equation}
Given these choices we can write the posterior probability of $k_i$  as
\begin{equation}
    \begin{aligned}
        p(k_i | x_i, y_i, \Theta) &= \frac{p(x_i, y_i |k_i, \Theta)p(k_i |\Theta)}{  p(x_i, y_i, k_i,\Theta)} \\
        & = \frac{p(y_i | x_i, k_i, \Theta)p(k_i, \Theta)p(x_i)}{p(x_i, y_i, k_i, \Theta) }\\
        & \propto p(y_i | x_i, k_i, \Theta) p(k_i| \Theta)p(x_i) \\
        & = p(y_i | q_i, x_i)p(q_i | k_i, \Theta)p(k_i | \Theta)p(x_i)\\
    \end{aligned}\label{eq:posterior}
\end{equation}
The model structure induced by these choices is illustrated in Figure~\ref{fig:model_dag}.

It remains to select the exact form of the given distributions.
The prior probability of being a delivery driver is simply a constant parameter, $p(k_i)=\theta$, and since $p(y \mid q,x)$ is the sum result of $x$ binary trials, each with probability $q$, it is given by the Binomial distribution, $p(y \mid q,x) = (n,x)p^{x}(1-p)^{n-x}$.
However, the form of $p(q \mid k)$ may be any distribution supported on $[0,1]$.
Selecting $p(q \mid k)$ as being drawn from a mixture of the Beta distributions,
\begin{equation}
p(q_i \mid k_i = k) = Beta(\alpha_k, \beta_k), k \in \{0, 1\}\label{eq:betaprior}
\end{equation}
is a natural choice as this is the conjugate prior to the Binomial distribution.
This is a well understood distribution with closed form expressions for mean and variance, allowing us some understanding of the prior distribution imposed on the form of $p(q)$ by our choice of hyperparameters.

To allow our hyperparameters to independently control our beliefs over the mean and a measure of the uncertainty of $p(q | k)$ we reparameterise our distributions such that our hyperparameters are given by
\begin{equation}
    \begin{aligned}
        \Phi &= \{\mu_0, \mu_1, r_0, r_1, \theta\} \\
        \mu_k &= \beta_k(\alpha_k + \beta_k) \\
        r_k &= \alpha_k + \beta_k - \tau_k \\
        \tau_k &= \max \left(\frac{\alpha_k + \beta_k}{\alpha_k}, \frac{\alpha_k + \beta_k}{\beta_k}\right) \\
    \end{aligned}\label{eq:hyperparameters}
\end{equation}
~(The offset term $\tau$ ensures that $\alpha_k$, $\beta_k > 0$ to avoid numerical issues).

\begin{figure}
    \centering
    \includegraphics[width=0.95\columnwidth]{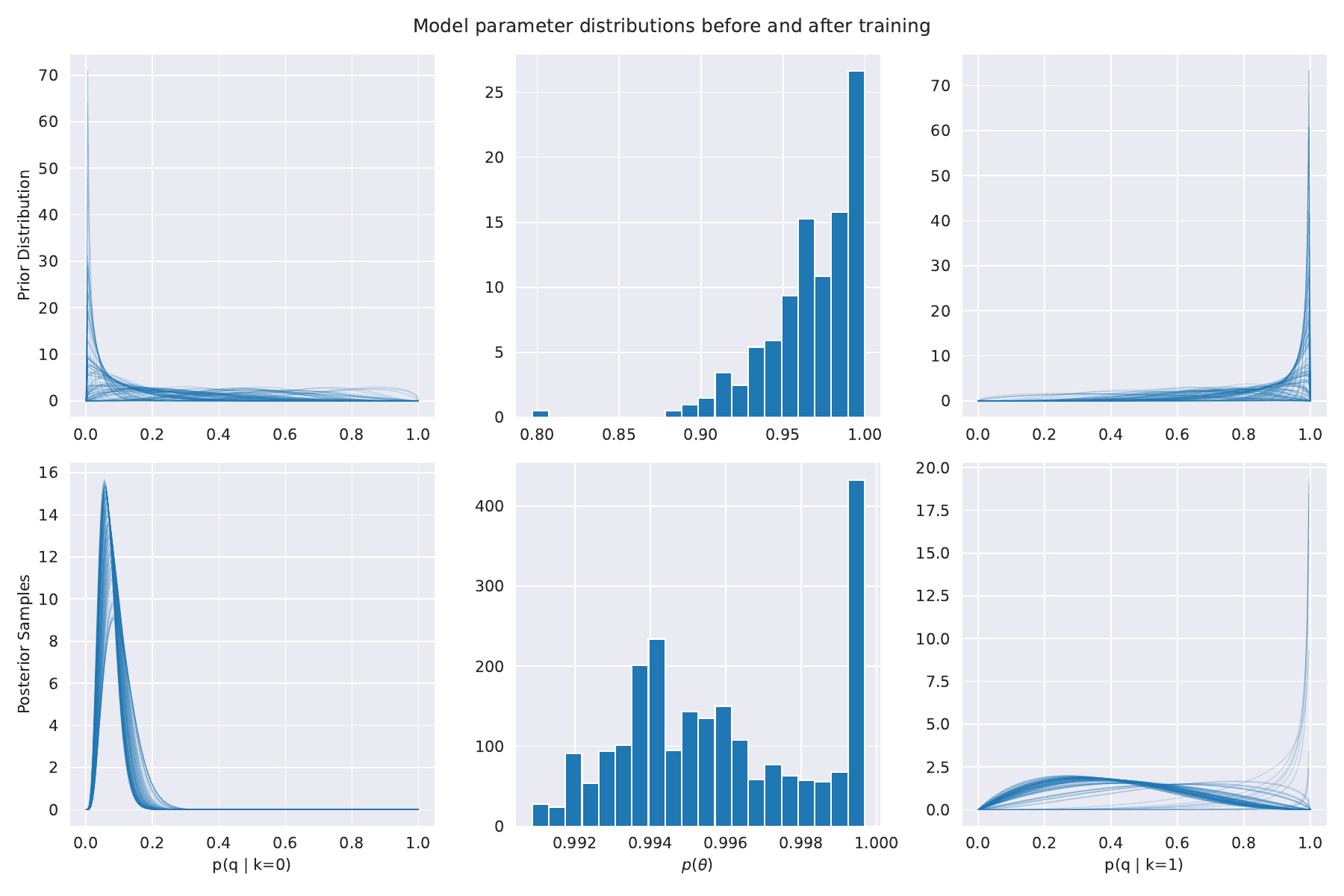}
    \caption{Samples of the prior (top row) and posterior (bottom row) of our belief over the distributions of $p(q \mid k = 0)$ (left column), $p(\theta)$ (centre column), and $p(q \mid k = 1)$ (right column).
    Under prior belief there is a relatively wide range of likely distributions, however after conditioning on observed data these distributions are more tightly defined.}
    \label{fig:mcmc-samples}
\end{figure}

In a setting in which we intended to identify unknown structure in the data we would select our hyperparameters, $\Phi$, to place broad uninformative priors over the model parameters $\Theta = \{\alpha_0, \alpha_1, \beta_0, \beta_1, \theta\}$.
However, in this setting we have relatively little data, and have strong belief that our classes are heavily imbalanced in favour of $k=0$  ($\theta$ is close to zero)
The majority class tends to have low rates of positive trip classification ($p(q_i \mid k=0)$ assigns most mass towards zero)
The minority class has much higher rates of positive trip classification  ($p(q_i \mid k=1)$ assigns most mass towards one).
We therefore choose more specific values
\begin{equation}
    \begin{aligned}
        p(\mu_0) &\sim Beta(1,3) \\
        p(\mu_1) &\sim Beta(4,1)\\
        p(r_0) &\sim Gamma(4,1)\\
        p(r_1) &\sim Gamma(8,1)\\
        p(\theta) &\sim Beta(30,1).
    \end{aligned}\label{eq:hyperpriors}
\end{equation}
The samples from the prior distributions induced over $p(\theta, q \mid k = 0)$ and $p(\theta, q \mid k = 1)$ are shown in the top row of Figure~\ref{fig:mcmc-samples}.

Given a probabilistic model equipped with priors over unknown parameters and an observed dataset, we have several available approaches to learn the parameters given the available data.
The final required posterior over class membership for each policy given training data, $p(k_i \mid x_i, y_i, X,Y,\Phi)$ is not available in closed form, so our inference method must be approximate.
We selected Markov-Chain Monte-Carlo (MCMC) inference as it allows us to specify informative priors, (allowing inference to be effective with limited data), as it provides a distribution over learned parameters (likely providing more uncertain predictions than  point estimate), and due to the availability of suitable implementations (we used a Hamiltonian Monte-Carlo implementation provided by the tensorflow-probability Python package~\cite{dillon2017tensorflow}).

Briefly, MCMC is an approach to drawing samples from a distribution given the ability to evaluate the likelihood (probability up to a constant factor) at a point.
Since the probability of the parameters conditioned on the data is proportional to the probability of the data conditioned on the parameters we can thus obtain samples of our model parameters from the joint distribution of model parameters conditioned on the data and hyperparameters $\Theta_j \sim p(\Theta | X,Y,\Phi)$.
For a proper introduction we recommend~\cite{murphy_machine_2012}.
From the resulting samples we can approximate the probability of class membership for any given policy given the associated trip counts
\begin{equation}
    \begin{aligned}
        p(k_i \mid x_i, y_i, \Phi, X,Y) &= \int_\Theta p(k_i \mid x_i, y_i, \Theta)p(\Theta \mid \Phi, X, Y) \mathrm{d}\Theta \\
        & \approx \frac{1}{N} \sum_j^N p(k_i \mid x_i, y_i, \Theta_j).
    \end{aligned}\label{eq:mcmcpost}
\end{equation}

While a posterior predictive probability of the minority class may be most useful in a machine learning setting it is not necessarily intuitive, or convenient to represent and compare values (many values will appear on screen as `0.9999\dots' or `0.0000\dots').
For this reason we pass the probability through a logit followed linear transform, such that a 99\% probability archives a score of 3, while the prior probability of minority class achieves a score of zero (Scores are clipped to remain between zero and 10).
As a monotonic transform this makes no difference to the ranking order, but is considered preferable for display purposes.

%% file: src/results.tex
Result gathering was performed in two phases.
In the first phase, following the initial modelling process detailed above, predictions from a test set of individual GPS traces were manually labelled by underwriters.
Following this the trained models were deployed on incoming data, with underwriters using the provided predictions to guide investigation, for which we present results gathered over a one year period.
Predictions and scores were updated on a weekly cadence, with a rolling 1-month window of data being included in predictions.

\subsection{Path Classification}\label{subsec:path-classification}
Following the development of the supervised model presented in Section~\ref{sec:gps-path-classification}, 100 delivery trip candidates were referred to the insurer’s Underwriting department, representing the 100 trips with the highest
predictive score as calculated by the model.
Out of these, 96 were confirmed to be true delivery trips.

Following this investigation, the model was deployed to make live predictions at the trip level.
In order to assess the policy-level model presented in Section~\ref{sec:policy-classification},
the trip-level model was run over GPS data corresponding to over 75000 distinct policies.

\subsection{Policy Classification}\label{subsec:policy-classification}
\begin{figure}
    \centering
    \includegraphics[width=0.95\columnwidth]{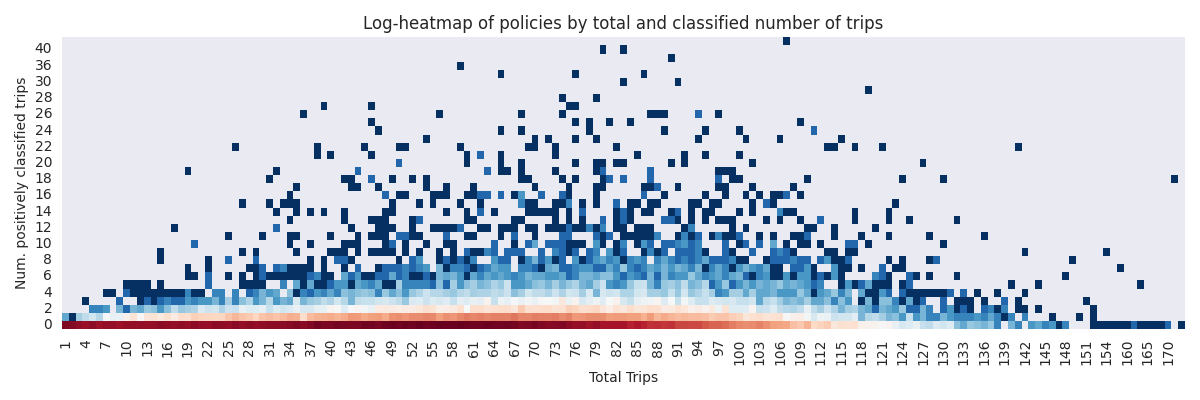}
    \caption{ Heatmap (hue is log-scaled) of policy counts by count of delivery and non-delivery classified trips (zero trips not included) While the overwhelming majority of policies have no positively classified trips, the number with a significant fraction still far exceeds manual processing capacity. Constructing a ranking of policies most likely to contain true-positive classifications allows efficient use of available human resources.}
    \label{fig:policy-heatmap}
\end{figure}

\begin{figure}
    \centering
    \includegraphics[width=0.7\columnwidth]{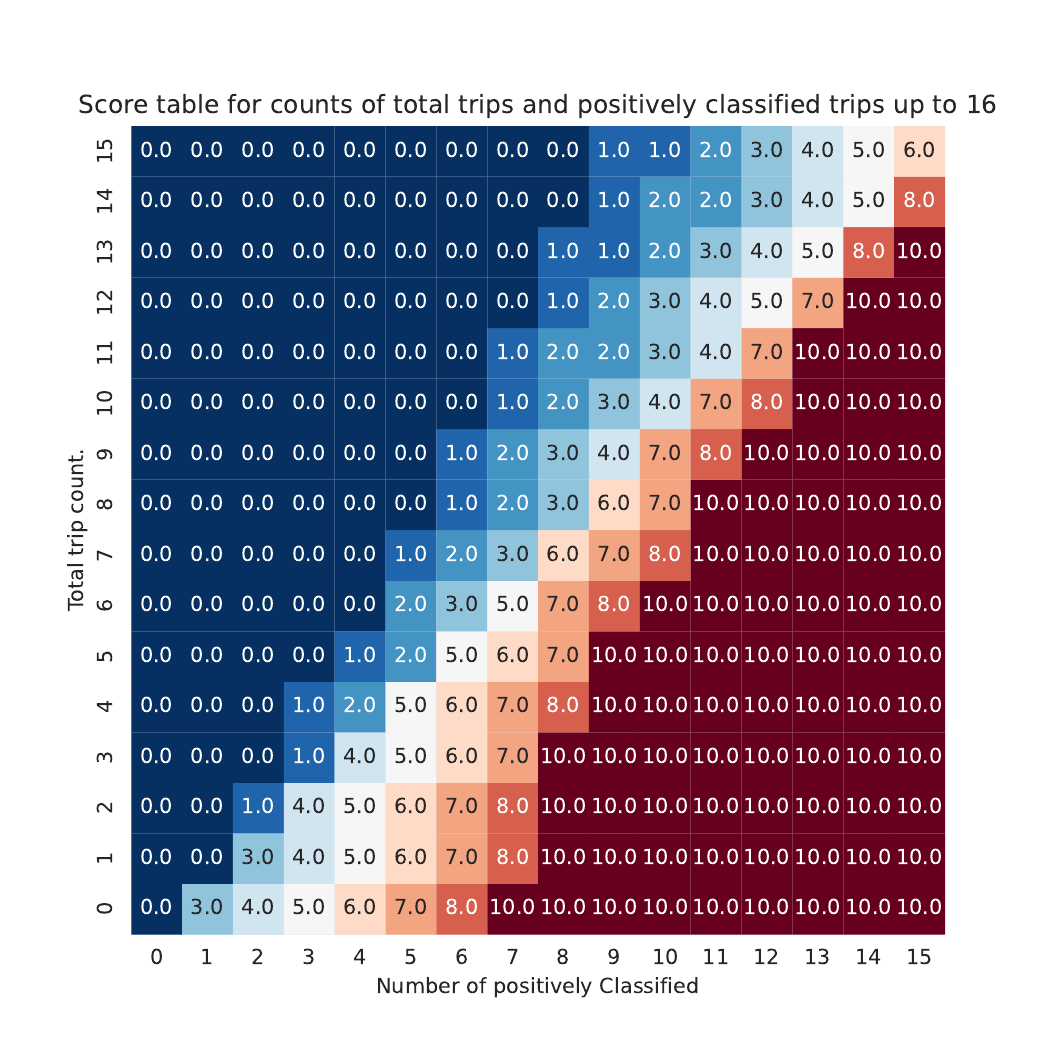}
    \caption{ Score lookup-table for total and positively classified trip counts.}
    \label{fig:score-table}
\end{figure}

At the time of training the policy classification model predictions were available for all trips undertaken in a one-month window, representing over 400000 trips undertaken by over 75000 distinct policies, the distribution of total and positively classified trips observed is illustrated in Figure~\ref{fig:policy-heatmap}.
Drawing 5000 parameters sampled from our MCMC model resulted in the posterior distributions shown in the lower row of Figure~\ref{fig:mcmc-samples}.
As illustrated this suggests that the mean rate of positively-classified trip generation by the majority (non-delivery) group is around 0.1, with the large majority being under 0.2, while for the minority class a much higher mean with significant probability of rates even up to 1.0 are predicted.
The posterior distribution on the probability of a policy being a member of the minority class given no observations is not predicted as likely to be greater than around 1\%, with a mean of 0.4\%.
From these parameters we can compute the probability of minority class membership for any number of positive and negatively classified individual trips, which is given in Figure~\ref{fig:score-table}.
We can see that this posterior map has the characteristics that we required.
For larger total numbers of trips a majority being positively identified represents a high likelihood of being a delivery driver, but for small numbers the fraction needs to be larger, and in particular policies having only 3 or 4 positively classified trips with no others are not considered conclusively to be delivery drivers.

\begin{figure}
    \centering
    \includegraphics[width=0.7\columnwidth]{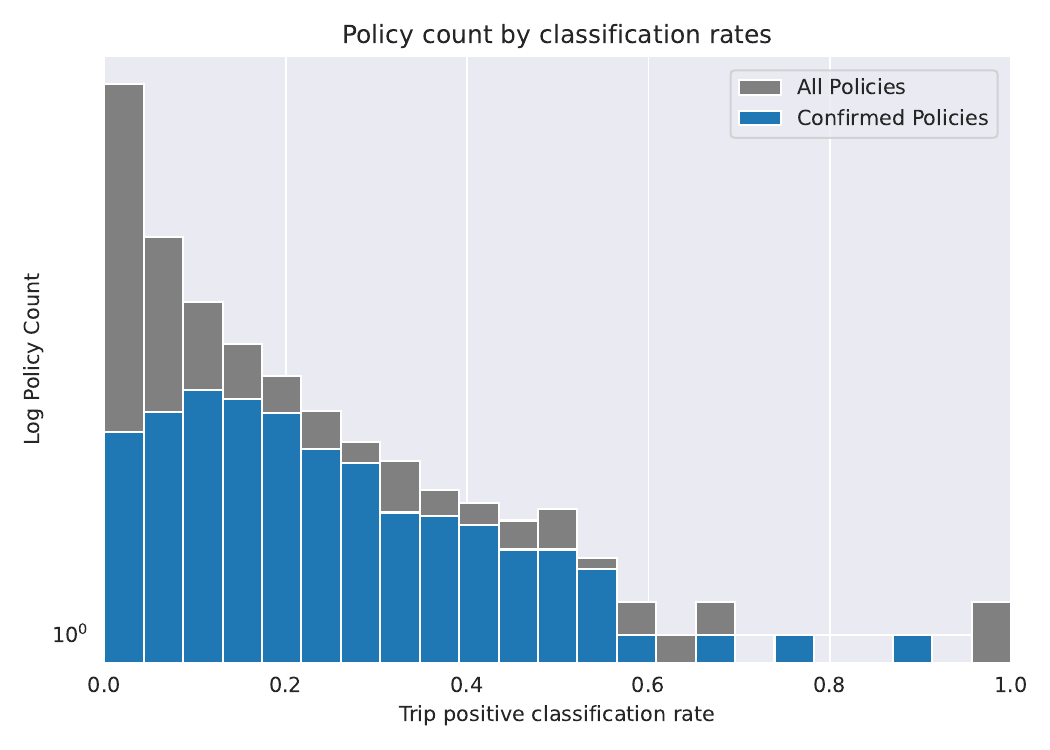}
    \caption{ Distribution of the fraction of trips being individually classified as delivery-driving overall and for identified and confirmed delivery driving polices (note the log scale). Policies that have made no trips are excluded.}
    \label{fig:count-histogram}
\end{figure}
During 1 year of trial use the scores and individual trip classifications were made available to underwriters investigating delivery driving.
Use of these quantities allowed highly scored policies, and positively classified trips undertaken by those policyholders to prioritise for investigation.
During this time approximately 0.8\% of individual policies were investigated (only 1 month of history is typically considered, so this represents at most 0.13\% of all trips and 2.6\% of individual positively classified trips being manually viewed due to machine learning predictions).
From these investigations only 0.4\% of reviewed policies were labelled by underwriters as not being delivery drivers (99.6\% accuracy).
This represents a highly efficient use of time due to machine learning, a random selection policy would be expected to provide a negative result for the overwhelming majority of policies, so is not a practical use of resources.

Successfully identifying a large number of policies engaging in delivery driving with minimal false positives demonstrates significant business value for the overall approach, which was originally designed on the basis of very limited positively labelled data.
The distribution of trips being individually classified for the overall population, and for confirmed delivery-driving policies after one year of data collection and manual investigation is shown in Figure~\ref{fig:count-histogram}.
The observed distributions are broadly consistent with those predicted using the modelling described in Section~\ref{sec:policy-classification}, and as noted above the predictions generated led to an accuracy of 99.6\% for policies that were given manual review, suggesting the approach is viable and provides useful predictions.
On the other hand, there are some notable differences between our modelling and the observed data.
First, the parameters generated by MCMC suggest a low prior probability of the minority class of approximately 0.4\%, with 1\% being in the tail of the distribution.
However, we have in fact obtained manual positive labels for 0.8\% of policies, and the minimal false positive rate for investigations suggests this may be a lower bound rather than the true number present.
It therefore seems likely that the true value of $\theta$ lies outside the range predicted by our modelling.
Second, our original assumption was that delivery drivers would incur a majority of trips being positively classified and this was encoded in our prior for $p(q \mid k=1)$ by setting the priors for $\mu_1$ and $r_1$ in Equation~\ref{eq:hyperpriors}.
However, both the learned posterior from initial training data, and the observed distribution after a year of data collection place the mean value of $p(q \mid k=1)$ at significantly less than $0.5$.
This has not in fact impacted the ability of the model to successfully make predictions, but does point to the importance of appropriate selection of modelling assumptions.

%% file: src/conclusions-and-future-work.tex
This work has detailed making use of machine learning to enable detection of driving characteristics,  automating a previously manual task which was of too great volume to be performed in full.
Automation ensures a consistent application of investigation, so enables fairer risk assessment and treatment of customers.
Provision of the resulting predictions in a ranked score enabled the insurer to focus on highest risk and manage risks appropriately with available resources.
Our approach has allowed a very restricted number of existing identified cases within the data to be expanded to positive identification of delivery driving in $\sim 0.8\%$ of policies on a rolling basis.
The process has provided a more effective and efficient means for the insurer to identify potential underinsurance or misrepresentation at scale, ultimately to improve business and customer outcome, and has provided an exemplar use case driving further adoption of human-AI collaboration within the business.
There is clear potential to expand the approach to other behaviours of interest to telematics insurers, including business use, private hire or other driving behaviour.
More general applications could also be explored, including classification problems where the approach could be adopted as part of a model to detect fraud, theft or other high risk events or measuring confidence around prediction in pricing / risk cost models.

%It is an example of Human and AI collaboration - fusion of experience design and quantitative methods that makes collaboration between humans and AI more intuitive, efficient and powerful.

From a modelling perspective initially only a limited number of confirmed cases of delivery driving had been identified, so training of the per-trip classifier was undertaken with very limited training data.
This resulted in a limited level of performance for the per-trip classification process.
Furthermore, the policy classification was undertaken in a fundamentally unsupervised manner, attempting to learn and distinguish two distinct but unlabelled classes from the available data.
Making use of these predictions to guide investigations has yielded significant new data which gives potential for significant further improvements.
In particular, following a year of guided investigations a much greater volume of labelled trips is available than at the original model training time.
This indicates the potential for further insights into possible feature engineering, and greater model performance on individual trip classification.
We now have positive labels for a significant number of policies undertaking delivery driving, and have learned that the distribution of positive trip-classification rates for delivery drivers is significantly different to our prior belief.
Updating our priors to reflect this new understanding, an increased volume of data, and including these partially available labels as observations of $y_i$ in the inference process can be expected to provide a much more effective policy classification metric in future.
It is also worth noting that any improvements to the individual trip classifier can be expected to yield greater separation, and therefore greater performance in policy classification.